\documentclass{article}
\usepackage[latin9]{luainputenc}
\usepackage{float}
\usepackage{amsmath}
\usepackage{graphicx}

\makeatletter

\floatstyle{ruled}
\newfloat{algorithm}{tbp}{loa}
\providecommand{\algorithmname}{Algorithm}
\floatname{algorithm}{\protect\algorithmname}

\usepackage{spconf}

\usepackage{amssymb}

\usepackage[english]{babel}
\usepackage{amsthm}
\newtheorem{theorem}{Theorem}
\newtheorem{lemma}[theorem]{Lemma}

\title{Robust Maximum Likelihood Estimation of Sparse Vector Error Correction Model}
\name{Ziping Zhao and Daniel P. Palomar\thanks{This work was supported by the Hong Kong RGC 16206315 research grant. Ziping Zhao is supported by the Hong Kong PhD Fellowship Scheme (HKPFS). (e-mail: ziping.zhao@connect.ust.hk; palomar@ust.hk).}}
\address{Department of Electronic and Computer Engineering,\\The Hong Kong University of Science and Technology, Hong Kong.}
\makeatother

\begin{document}
\maketitle 
\begin{abstract}
In econometrics and finance, the vector error correction model (VECM)
is an important time series model for cointegration analysis, which
is used to estimate the long-run equilibrium variable relationships.
The traditional analysis and estimation methodologies assume the underlying
Gaussian distribution but, in practice, heavy-tailed data and outliers
can lead to the inapplicability of these methods. In this paper, we
propose a robust model estimation method based on the Cauchy distribution
to tackle this issue. In addition, sparse cointegration relations
are considered to realize feature selection and dimension reduction.
An efficient algorithm based on the majorization-minimization (MM)
method is applied to solve the proposed nonconvex problem. The performance
of this algorithm is shown through numerical simulations.
\end{abstract}

\section{Introduction}

The vector error correction model (VECM) \cite{Luetkepohl2007} is
very important in cointegration analysis to estimate and test for
the long-run cointegrated equilibriums. It is widely used in time
series modeling for financial returns and macroeconomic variables.
In \cite{Granger1983,EngleGranger1987}, Engle and Granger first proposed
the concept of ``cointegration'' to describe the linear stationary
relationships in the nonstationary time series. Later, Johansen studied
the statistical estimation and inference problem in time series cointegration
modeling \cite{Johansen1988,Johansen1991,Johansen1995a}. A VECM for
$\mathbf{y}_{t}\in\mathbb{R}^{K}$ is given as follows:
\begin{equation}
\begin{array}{c}
\Delta\mathbf{y}_{t}=\boldsymbol{\nu}+\boldsymbol{\Pi}\mathbf{y}_{t-1}+\sum_{i=1}^{p-1}\boldsymbol{\Gamma}_{i}\Delta\mathbf{y}_{t-i}+\boldsymbol{\varepsilon}_{t},\end{array}\label{eq:VECM}
\end{equation}
where $\Delta$ is the first difference operator, i.e., $\Delta\mathbf{y}_{t}=\mathbf{y}_{t}-\mathbf{y}_{t-1}$,
$\boldsymbol{\nu}$ denotes the drift, $\boldsymbol{\Pi}$ determines
the long-run equilibriums, $\boldsymbol{\Gamma}_{i}$ ($i=1,\ldots,p-1$)
contains the short-run effects, and $\boldsymbol{\varepsilon}_{t}$
is the innovation with mean $\mathbf{0}$ and covariance $\boldsymbol{\Sigma}$.
Matrix $\boldsymbol{\Pi}$ has a reduced cointegration rank $r$,
i.e., $\mathrm{rank}\left(\boldsymbol{\Pi}\right)=r<K$, and it can
be written as $\boldsymbol{\Pi}=\boldsymbol{\alpha}\boldsymbol{\beta}^{T}$
($\boldsymbol{\alpha},\boldsymbol{\beta}\in\mathbb{R}^{K\times r}$).
Accordingly, $\mathbf{y}_{t}$ is said to be cointegrated with rank
$r$, and $\boldsymbol{\beta}^{T}\mathbf{y}_{t}$ gives the long-run
stationary time series defined by the cointegration matrix $\boldsymbol{\beta}$.
Such long-run equilibriums are often implied by economic theory and
can be used for statistical arbitrage \cite{Pole2011}.

It is well-known that financial returns and macroeconomic variables
exhibit heavy-tails and are often associated with outliers due to
external factors, like political and regulatory changes, as well as
data corruption, like faulty observations and wrongly processed data
\cite{RachevMennFabozzi2005}. These stylized features contradict
the popular Gaussian noise assumption typically made in the theoretical
analysis and estimation procedures with adverse effects in the estimated
models. Cointegration analysis is particularly sensitive to these
issues. Papers \cite{FransesHaldrup1994,FransesKloekLucas1998,BohnNielsen2004}
discussed the properties of the Dickey-Fuller test and the Johansen
test in the presence of outliers. Lucas studied such issues both from
a theoretical and an empirical point of view \cite{Lucas1995,Lucas1995a,FransesLucas1998}.
To deal with the heavy-tails and outliers in time series modeling,
simple and effective estimation methods are needed. In \cite{Lucas1997},
the pseudo maximum likelihood estimators were introduced for VECM.
In this paper, based on \cite{Lucas1997}, we formulate the estimation
problem based on the log-likelihood function of the Cauchy distribution
as a conservative representative of the heavy-tailed distributions
to better fit the heavy-tails and dampen the influence of outliers.

Sparse optimization \cite{BachJenattonMairalObozinskiothers2012}
has become the focus of much research interest as a way to realize
feature selection and dimension reduction (e.g., lasso \cite{Tibshirani1996}).
In \cite{WilmsCroux2016}, element-wise sparsity was imposed on $\boldsymbol{\beta}$
in VECM modeling. As indicated by \cite{ChenHuang2012,ChenHuang2016},
to realize the feature selection purpose, group sparsity is better
since it can simultaneously reduce the same variable in all cointegration
relations and naturally keep the geometry of the low-rank parameter
space. In this paper, instead of imposing the group sparsity on $\boldsymbol{\beta}$,
we equivalently put group sparsity on $\boldsymbol{\Pi}$ and add
a rank constraint for it, which can realize the same target without
the ahead factorization $\boldsymbol{\Pi}=\boldsymbol{\alpha}\boldsymbol{\beta}^{T}$.
For sparsity pursuing, i.e., approximating the $\ell_{0}$-``norm'',
rather than the popular $\ell_{1}$-norm, we use a nonconvex Geman-type
function \cite{GemanReynolds1992} which has a better approximation
power. A smoothed counterpart is also firstly proposed to reduce the
``singularity issue'' in optimization, based on which the group
sparsity regularizer of $\boldsymbol{\beta}$ is attained.

Robust estimation is somewhat underrated in financial applications
due to the complex computations that are time and resource intensive.
By considering the robust loss and the regularizer, a nonconvex optimization
problem is finally formulated. The expectation-maximization (EM) is
usually used to solve the robust losses (e.g., \cite{BoscoParisioPelagattiBaldi2010}).
However, EM cannot be applied for our formulation. To deal with it,
an efficient algorithm based on the majorization-minimization (MM)
method is proposed with estimation performance numerically shown. 

\section{Robust Estimation of Sparse VECM\label{sec:Robust-Estimation-of-VECM}}

Suppose a sample path $\left\{ \mathbf{y}_{t}\right\} _{t=1}^{N}$
($N>K$) and the needed pre-sample values are available, then the
VECM \eqref{eq:VECM} can be written into a matrix form as follows:
\begin{equation}
\begin{aligned}\Delta\mathbf{Y} & =\boldsymbol{\Pi}\mathbf{Y}_{-1}+\boldsymbol{\Gamma}\Delta\mathbf{X}+\mathbf{E},\end{aligned}
\label{eq:VECM sample}
\end{equation}
where $\boldsymbol{\Gamma}=\left[\boldsymbol{\Gamma}_{1},\ldots,\boldsymbol{\Gamma}_{p-1},\boldsymbol{\nu}\right]$,
$\Delta\mathbf{Y}=\left[\Delta\mathbf{y}_{1},\ldots,\Delta\mathbf{y}_{N}\right]$,
$\mathbf{Y}_{-1}=\left[\mathbf{y}_{0},\ldots,\mathbf{y}_{N-1}\right]$,
$\Delta\mathbf{X}=\left[\Delta\mathbf{x}_{1},\ldots,\Delta\mathbf{x}_{N}\right]$
with $\Delta\mathbf{x}_{t}=\left[\begin{array}{c}
\Delta\mathbf{y}_{t-1}^{T},\ldots,\Delta\mathbf{y}_{t-p+1}^{T},1\end{array}\right]^{T}$, and $\mathbf{E}=\left[\boldsymbol{\varepsilon}_{1},\ldots,\boldsymbol{\varepsilon}_{N}\right]$. 

\subsection{Robustness Pursued by Cauchy Log-likelihood Loss}

The robustness is pursued by a multivariate Cauchy distribution. Assume
the innovations $\boldsymbol{\varepsilon}_{t}$'s in \eqref{eq:VECM}
follow Cauchy distribution, i.e., $\boldsymbol{\varepsilon}_{t}\sim\mathrm{Cauchy}\left(\boldsymbol{0},\boldsymbol{\Sigma}\right)$
with $\boldsymbol{\Sigma}\in\mathbb{S}_{++}^{K}$, then the probability
density function is given by
\[
g_{\boldsymbol{\theta}}\left(\boldsymbol{\varepsilon}_{t}\right)=\frac{\Gamma\left(\frac{1+K}{2}\right)}{\Gamma\left(\frac{1}{2}\right)\left(\nu\pi\right)^{\frac{K}{2}}}\left[\det\left(\boldsymbol{\Sigma}\right)\right]^{-\frac{1}{2}}\left(1+\boldsymbol{\varepsilon}_{t}^{T}\boldsymbol{\Sigma}^{-1}\boldsymbol{\varepsilon}_{t}\right)^{-\frac{1+K}{2}}.
\]
The negative log-likelihood loss function of the Cauchy distribution
for $N$ samples from \eqref{eq:VECM} is written as follows:
\begin{equation}
\begin{array}{rl}
 & L\left(\boldsymbol{\theta}\right)=\frac{N}{2}\log\det\left(\boldsymbol{\Sigma}\right)+\frac{1+K}{2}\sum_{i=1}^{N}\log\Bigl(1+\\
 & \quad\quad\quad\left\Vert \boldsymbol{\Sigma}^{-\frac{1}{2}}\left(\Delta\mathbf{y}_{i}-\boldsymbol{\Pi}\mathbf{y}_{i-1}-\boldsymbol{\Gamma}\Delta\mathbf{x}_{i-1}\right)\right\Vert _{2}^{2}\Bigr),
\end{array}\label{eq:loss}
\end{equation}
where the constants are dropped and $\boldsymbol{\theta}\triangleq\left\{ \boldsymbol{\Pi}\left(\boldsymbol{\alpha},\boldsymbol{\beta}\right),\boldsymbol{\Gamma},\boldsymbol{\Sigma}\right\} $.

\subsection{Group Sparsity Pursued by Nonconvex Regularizer}

For a vector $\mathbf{x}\in\mathbb{R}^{K}$, the sparsity level is
usually measured by the $\ell_{0}$-``norm'' (or $\mathrm{sgn}\left(\left|x\right|\right)$)
as $\left\Vert \mathbf{x}\right\Vert _{0}=\sum_{i=1}^{K}\mathrm{sgn}\left(\left|x_{i}\right|\right)=k$,
where $k$ is the number of nonzero entries in $\mathbf{x}$. Generally,
applying the $\ell_{0}$-``norm'' to different groups of variables
can enforce group sparsity in the solutions. The $\ell_{0}$-``norm''
is not convex and not continuous, which makes it computationally difficult
and leads to intractable NP-hard problems. So, $\ell_{1}$-norm as
the tightest convex relaxation is usually used to approximate the
$\ell_{0}$-``norm'' in practice, which is easier for optimization
and still favors sparse solutions.

Tighter nonconvex sparsity-inducing functions can lead to better performance
\cite{BachJenattonMairalObozinskiothers2012}. In this paper, to better
pursue the sparsity and to remove the ``singularity issue'', i.e.,
when using nonsmooth functions, the variable may get stuck at a nonsmooth
point \cite{FigueiredoBioucas-DiasNowak2007}, a smooth nonconvex
function based on the rational (Geman) function in \cite{GemanReynolds1992}
is used given as follows:
\[
\begin{array}{l}
\mathrm{rat}_{p}^{\epsilon}\left(x\right)=\begin{cases}
\frac{px^{2}}{2\epsilon\left(p+\epsilon\right)^{2}}, & \left|x\right|\leq\epsilon\\
\frac{\left|x\right|}{p+\left|x\right|}-\frac{2\epsilon^{2}+p\epsilon}{2\left(p+\epsilon\right)^{2}}, & \left|x\right|>\epsilon
\end{cases}.\end{array}
\]
In order to attain feature selection in VECM, i.e., sparse cointegration
relations, according to \cite{ChenHuang2012,ChenHuang2016}, we can
impose the row-wise group sparsity on matrix $\boldsymbol{\beta}$.
In fact, due to $\boldsymbol{\Pi}=\boldsymbol{\alpha}\boldsymbol{\beta}^{T}$,
the row-wise sparsity imposed on $\boldsymbol{\beta}$ can also be
realized by directly estimating $\boldsymbol{\Pi}$ through imposing
the column-wise group sparsity on $\boldsymbol{\Pi}$ and constraining
its rank. Then we have the sparsity regularizer of matrix $\boldsymbol{\Pi}$
which is given by
\begin{equation}
\begin{array}{c}
R\left(\boldsymbol{\Pi}\right)=\sum_{i=1}^{K}\mathrm{rat}_{p}^{\epsilon}\left(\left\Vert \boldsymbol{\pi}_{i}\right\Vert _{2}\right),\end{array}\label{eq:regularization}
\end{equation}
where $\boldsymbol{\pi}_{i}$ ($i=1,\ldots,K$) denotes the $i$th
column of $\boldsymbol{\Pi}$. The grouping effect is achieved by
taking the $\ell_{2}$-norm of each group, and then applying the group
regularization.

\subsection{Problem Formulation}

By combining the robust loss function \eqref{eq:loss} and the sparsity
regularizer \eqref{eq:regularization}, we attain a penalized maximum
likelihood estimation formulation which is specified as follows:
\begin{equation}
\begin{array}{cl}
\underset{\boldsymbol{\theta}=\left\{ \boldsymbol{\Pi},\boldsymbol{\Gamma},\boldsymbol{\Sigma}\right\} }{\mathsf{minimize}} & F\left(\boldsymbol{\theta}\right)\triangleq L\left(\boldsymbol{\theta}\right)+\xi R\left(\boldsymbol{\Pi}\right)\\
\mathsf{subject\:to} & \mathrm{rank}\left(\boldsymbol{\Pi}\right)\leq r,\:\boldsymbol{\Sigma}\succeq\mathbf{0}.
\end{array}\label{eq:Problem Formulation}
\end{equation}
This is a constrained smooth nonconvex problem due to the nonconvexity
of the objective function and the constraint set.

\section{Problem Solving via The MM Method\label{sec:Problem-Solving}}

The MM method \cite{HunterLange2004,RazaviyaynHongLuo2013,SunBabuPalomar2016}
is a generalization of the well-known EM method. For an optimization
problem given by
\[
\begin{array}{lclc}
\underset{\mathbf{x}}{\mathsf{minimize}} & f\left(\mathbf{x}\right) & \mathsf{subject\:to} & \mathbf{x}\in{\cal X},\end{array}
\]
instead of dealing with this problem directly which could be difficult,
the MM-based algorithm solves a series of simpler subproblems with
surrogate functions that majorize $f\left(\mathbf{x}\right)$ over
${\cal X}$. More specifically, starting from an initial point $\mathbf{x}^{\left(0\right)}$,
it produces a sequence $\left\{ \mathbf{x}^{\left(k\right)}\right\} $
by the following update rule:
\[
\mathbf{x}^{\left(k\right)}\in\arg\min_{\mathbf{x}\in{\cal X}}\:\overline{f}\left(\mathbf{x},\mathbf{x}^{\left(k-1\right)}\right),
\]
where the surrogate majorizing function $\overline{f}\left(\mathbf{x},\mathbf{x}^{\left(k\right)}\right)$
satisfies
\[
\begin{array}{cl}
\overline{f}\left(\mathbf{x}^{\left(k\right)},\mathbf{x}^{\left(k\right)}\right)=f\left(\mathbf{x}^{\left(k\right)}\right), & \forall\mathbf{x}^{\left(k\right)}\in{\cal X},\\
\overline{f}\left(\mathbf{x},\mathbf{x}^{\left(k\right)}\right)\geq f\left(\mathbf{x}\right), & \forall\mathbf{x},\mathbf{x}^{\left(k\right)}\in{\cal X},\\
\overline{f}^{\prime}\left(\mathbf{x}^{\left(k\right)},\mathbf{x}^{\left(k\right)};\mathbf{d}\right)=f^{\prime}\left(\mathbf{x}^{\left(k\right)};\mathbf{d}\right), & \forall\mathbf{d},\mbox{\text{ s.t. }}\mathbf{x}^{\left(k\right)}+\mathbf{d}\in{\cal X}.
\end{array}
\]
The objective function value is monotonically nonincreasing at each
iteration. In order to use the MM method, the key step is to find
a majorizing function to make the subproblem easy to solve, which
will be discussed in the following subsections.

\subsection{Majorization for the Robust Loss Function $L\left(\boldsymbol{\theta}\right)$}

Instead of using the EM method \cite{BoscoParisioPelagattiBaldi2010},
in this paper, we derive the majorizing function for $L\left(\boldsymbol{\theta}\right)$
from an MM perspective.
\begin{lemma}\label{lem:log mahorization}At any point $x^{\left(k\right)}\in\mathbb{R}$,
$\log\left(1+x\right)\leq\log\left(1+x^{\left(k\right)}\right)$ $+\frac{1}{1+x^{\left(k\right)}}\left(x-x^{\left(k\right)}\right)$,
with the equality attained at $x=x^{\left(k\right)}$.\end{lemma}
Based on Lemma \ref{lem:log mahorization}, at the iterate $\boldsymbol{\theta}^{\left(k\right)}$,
the loss function $L\left(\boldsymbol{\theta}\right)$ can be majorized
by the following function:

\vspace{-10bp}
\[
\begin{array}{l}
\overline{L}_{1}\left(\boldsymbol{\theta},\boldsymbol{\theta}^{\left(k\right)}\right)\simeq\\
\frac{N}{2}\log\det\left(\boldsymbol{\Sigma}\right)+\frac{1}{2}\left\Vert \boldsymbol{\Sigma}^{-\frac{1}{2}}\left(\Delta\bar{\mathbf{Y}}-\boldsymbol{\Pi}\bar{\mathbf{Y}}_{-1}-\boldsymbol{\Gamma}\Delta\bar{\mathbf{X}}\right)\right\Vert _{F}^{2},
\end{array}
\]
where ``$\simeq$'' means ``equivalence'' up to additive constants,
$\Delta\bar{\mathbf{Y}}=\Delta\mathbf{Y}\mathrm{diag}\left(\sqrt{\mathbf{w}^{\left(k\right)}}\right)$,
$\bar{\mathbf{Y}}_{-1}=\mathbf{Y}_{-1}\mathrm{diag}\left(\sqrt{\mathbf{w}^{\left(k\right)}}\right)$,
and $\Delta\bar{\mathbf{X}}=\Delta\mathbf{X}\mathrm{diag}\left(\sqrt{\mathbf{w}^{\left(k\right)}}\right)$
with $\mathbf{w}^{\left(k\right)}\in\mathbb{R}^{N}$ and the element
\[
\begin{array}{c}
w_{t}^{\left(k\right)}=\frac{1+K}{1+\left\Vert \boldsymbol{\Sigma}^{-\frac{\left(k\right)}{2}}\left(\Delta\mathbf{y}_{t}-\boldsymbol{\Pi}^{\left(k\right)}\mathbf{y}_{t-1}-\boldsymbol{\Gamma}^{\left(k\right)}\Delta\mathbf{x}_{t-1}\right)\right\Vert _{2}^{2}},\;t=1\ldots N.\end{array}
\]
By taking the partial derivatives for $\boldsymbol{\Sigma}$ and $\boldsymbol{\Gamma}$,
and defining the projection matrix $\bar{\mathbf{M}}=\mathbf{I}_{N}-\Delta\bar{\mathbf{X}}^{T}\left(\Delta\bar{\mathbf{X}}\Delta\bar{\mathbf{X}}^{T}\right)^{-1}\Delta\bar{\mathbf{X}}$,
the majorizing function $\overline{L}_{1}\left(\boldsymbol{\theta},\boldsymbol{\theta}^{\left(k\right)}\right)$
is minimized when\vspace{-8bp}
\[
\begin{array}{l}
\boldsymbol{\Gamma}\left(\boldsymbol{\Pi}\right)=\left(\Delta\bar{\mathbf{Y}}-\boldsymbol{\Pi}\bar{\mathbf{Y}}_{-1}\right)\Delta\bar{\mathbf{X}}^{T}\left(\Delta\bar{\mathbf{X}}\Delta\bar{\mathbf{X}}^{T}\right)^{-1},\\
\boldsymbol{\Sigma}\left(\boldsymbol{\Pi}\right)=\frac{1}{N}\left(\Delta\bar{\mathbf{Y}}-\boldsymbol{\Pi}\bar{\mathbf{Y}}_{-1}\right)\bar{\mathbf{M}}\left(\Delta\bar{\mathbf{Y}}-\boldsymbol{\Pi}\bar{\mathbf{Y}}_{-1}\right)^{T}.
\end{array}
\]
Substituting these equations back into $\overline{L}_{1}\left(\boldsymbol{\theta},\boldsymbol{\theta}^{\left(k\right)}\right)$,
we have\vspace{-8bp}
\[
\begin{array}{l}
\overline{L}_{1}\left(\boldsymbol{\Pi},\boldsymbol{\theta}^{\left(k\right)}\right)\simeq\\
\frac{N}{2}\log\det\left[\left(\Delta\bar{\mathbf{Y}}-\boldsymbol{\Pi}\bar{\mathbf{Y}}_{-1}\right)\bar{\mathbf{M}}\left(\Delta\bar{\mathbf{Y}}-\boldsymbol{\Pi}\bar{\mathbf{Y}}_{-1}\right)^{T}\right].
\end{array}
\]
Then we introduce the following useful lemma.
\begin{lemma}\label{lem:logdet majorization}At any point $\mathbf{R}^{\left(k\right)}\in\mathbb{S}_{++}^{K}$,
$\log\det\left(\mathbf{R}\right)\leq\mathrm{Tr}\left(\mathbf{R}^{-\left(k\right)}\mathbf{R}\right)+\log\det\left(\mathbf{R}^{\left(k\right)}\right)-K$,
with the equality attained at $\mathbf{R}=\mathbf{R}^{\left(k\right)}$.\end{lemma}
Based on Lemma \ref{lem:logdet majorization}, $\overline{L}_{1}\left(\boldsymbol{\Pi},\boldsymbol{\theta}^{\left(k\right)}\right)$
is further majorized by
\[
\begin{array}{c}
\overline{L}_{2}\left(\boldsymbol{\Pi},\boldsymbol{\theta}^{\left(k\right)}\right)\simeq\frac{1}{2}\left\Vert \boldsymbol{\Sigma}^{-\frac{\left(k\right)}{2}}\left(\Delta\bar{\mathbf{Y}}-\boldsymbol{\Pi}\bar{\mathbf{Y}}_{-1}\right)\bar{\mathbf{M}}\right\Vert _{F}^{2}\end{array}.
\]
Finally, after majorization, $\overline{L}_{2}\left(\boldsymbol{\Pi},\boldsymbol{\theta}^{\left(k\right)}\right)$
becomes a quadratic function in $\boldsymbol{\Pi}$.\vspace{-10bp}

\subsection{Majorization for the Sparsity Regularizer $R\left(\boldsymbol{\Pi}\right)$}

\begin{figure}[tb]
\centering{}\includegraphics[bb=44bp 23bp 430bp 404bp,scale=0.46]{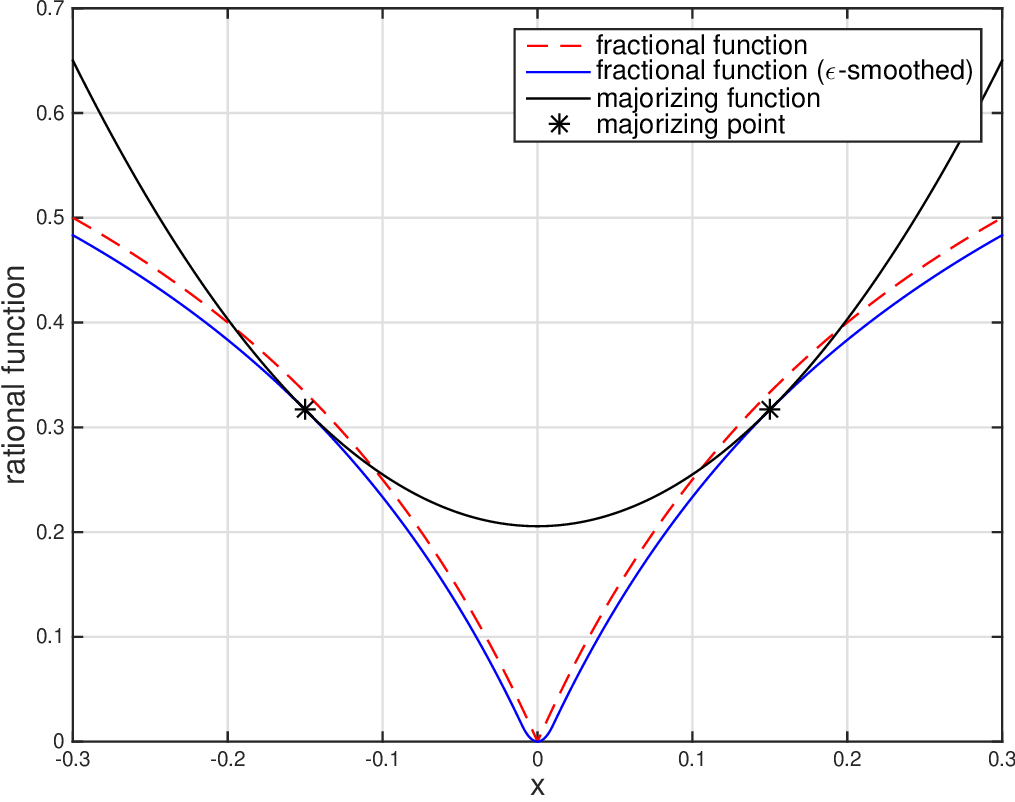}\caption{\label{fig:Majorization-for-the-sparsity}Majorization for smoothed
sparsity-inducing function.}
\end{figure}

In this section, we introduce the majorization trick to deal with
the nonconvex sparsity regularizer $R\left(\boldsymbol{\Pi}\right)$.
\begin{lemma}\label{lem:Sparsity quadratic}At any given point $x^{\left(k\right)}$,
$\mathrm{rat}_{p}^{\epsilon}\left(x\right)\leq\frac{q^{\left(k\right)}}{2}x^{2}+c^{\left(k\right)}$,
with the equality attained at $x=x^{\left(k\right)}$, the coefficient
$\begin{array}{c}
q^{\left(k\right)}=p\left[\max\left\{ \epsilon,\left|x^{\left(k\right)}\right|\right\} \left(p+\max\left\{ \epsilon,\left|x^{\left(k\right)}\right|\right\} \right)^{2}\right]^{-1}\end{array}$, and constant $\begin{array}{c}
c^{\left(k\right)}=\frac{p\max\left\{ \epsilon,\left|x^{\left(k\right)}\right|\right\} +2\left(\max\left\{ \epsilon,\left|x^{\left(k\right)}\right|\right\} \right)^{2}}{2\left(p+\max\left\{ \epsilon,\left|x^{\left(k\right)}\right|\right\} \right)^{2}}-\frac{p\epsilon+2\epsilon^{2}}{2\left(p+\epsilon\right)^{2}}\end{array}$.\end{lemma}
The majorization in Lemma \ref{lem:Sparsity quadratic} is pictorially
shown in Fig. \ref{fig:Majorization-for-the-sparsity}. Then at $\boldsymbol{\theta}^{\left(k\right)}$,
the regularizer $R\left(\boldsymbol{\Pi}\right)$ can be majorized
by
\[
\begin{array}{c}
\overline{R}\left(\boldsymbol{\Pi},\boldsymbol{\theta}^{\left(k\right)}\right)\simeq\frac{1}{2}\mathrm{vec}\left(\boldsymbol{\Pi}\right)^{T}\left[\mathrm{diag}\left(\mathbf{q}^{\left(k\right)}\right)\otimes\mathbf{I}_{K}\right]\mathrm{vec}\left(\boldsymbol{\Pi}\right),\end{array}
\]
where $\mathbf{q}^{\left(k\right)}\in\mathbb{R}^{K}$ with the $i$th
($i=1,\ldots,K$) element\vspace{-8bp}
\[
\begin{array}{c}
q_{i}^{\left(k\right)}=p\left[\max\left\{ \epsilon,\left\Vert \boldsymbol{\pi}_{i}^{\left(k\right)}\right\Vert _{2}\right\} \left(p+\max\left\{ \epsilon,\left\Vert \boldsymbol{\pi}_{i}^{\left(k\right)}\right\Vert _{2}\right\} \right)^{2}\right]^{-1}.\end{array}
\]

\vspace{-20bp}

\subsection{The Majorized Subproblem in MM}

By combining $\overline{L}_{2}\left(\boldsymbol{\Pi},\boldsymbol{\theta}^{\left(k\right)}\right)$
and $\overline{R}\left(\boldsymbol{\Pi},\boldsymbol{\theta}^{\left(k\right)}\right)$,
we can get the majorizing function for $\overline{F}\left(\boldsymbol{\theta}\right)$
which is given as follows:\vspace{-5bp}
\[
\begin{array}{l}
\overline{F}_{1}\left(\boldsymbol{\Pi},\boldsymbol{\theta}^{\left(k\right)}\right)\simeq\overline{L}_{2}\left(\boldsymbol{\Pi},\boldsymbol{\theta}^{\left(k\right)}\right)+\xi\overline{R}\left(\boldsymbol{\Pi},\boldsymbol{\theta}^{\left(k\right)}\right)\\
\simeq\frac{1}{2}\mathrm{vec}\left(\boldsymbol{\Pi}\right)^{T}\mathbf{G}^{\left(k\right)}\mathrm{vec}\left(\boldsymbol{\Pi}\right)-\mathrm{vec}\left(\mathbf{H}^{\left(k\right)}\right)^{T}\mathrm{vec}\left(\boldsymbol{\Pi}\right),
\end{array}
\]
where $\mathbf{G}^{\left(k\right)}=\bar{\mathbf{Y}}_{-1}\bar{\mathbf{M}}\bar{\mathbf{Y}}_{-1}^{T}\otimes\boldsymbol{\Sigma}^{-\left(k\right)}+\xi\mathrm{diag}\left(\mathbf{q}^{\left(k\right)}\right)\otimes\mathbf{I}_{K}$,
and $\mathbf{H}^{\left(k\right)}=\boldsymbol{\Sigma}^{-\left(k\right)}\Delta\bar{\mathbf{Y}}\bar{\mathbf{M}}\bar{\mathbf{Y}}_{-1}^{T}$.
Although $\overline{F}_{1}\left(\boldsymbol{\Pi},\boldsymbol{\theta}^{\left(k\right)}\right)$
is a quadratic function in $\boldsymbol{\Pi}$, together with the
nonconvex rank constraint on $\boldsymbol{\Pi}$ in \eqref{eq:Problem Formulation},
the problem is still hard to solve.
\begin{lemma}\label{lem:quadratic majorization}Let $\mathbf{A},\mathbf{B}\in\mathbb{S}^{K}$
and $\mathbf{B}\succeq\mathbf{A}$, then at any point $\mathbf{x}^{\left(k\right)}\in\mathbb{R}^{K}$,
$\mathbf{x}^{T}\mathbf{A}\mathbf{x}\leq\mathbf{x}^{T}\mathbf{B}\mathbf{x}+2\mathbf{x}^{\left(k\right)T}\left(\mathbf{A}-\mathbf{B}\right)\mathbf{x}+\mathbf{x}^{\left(k\right)T}\left(\mathbf{B}-\mathbf{A}\right)\mathbf{x}^{\left(k\right)}$
with the equality attained at $\mathbf{x}=\mathbf{x}^{\left(k\right)}$.\end{lemma}
Based on Lemma \ref{lem:quadratic majorization} and noticing $\psi_{\mathbf{G}}^{\left(k\right)}\mathbf{I}_{K^{2}}\succeq\mathbf{G}^{\left(k\right)}$
for any $\psi_{\mathbf{G}}^{\left(k\right)}$ satisfying $\psi_{\mathbf{G}}^{\left(k\right)}\geq\lambda_{\max}\left(\mathbf{G}^{\left(k\right)}\right)$,
$\overline{F}_{1}\left(\boldsymbol{\Pi},\boldsymbol{\theta}^{\left(k\right)}\right)$
can be further majorized by the following function:\vspace{-5bp}
\[
\begin{array}{l}
\overline{F}_{2}\left(\boldsymbol{\Pi},\boldsymbol{\theta}^{\left(k\right)}\right)\simeq\frac{1}{2}\psi_{\mathbf{G}}^{\left(k\right)}\left\Vert \boldsymbol{\Pi}-\mathbf{P}^{\left(k\right)}\right\Vert _{F}^{2},\end{array}
\]
where $\mathbf{P}^{\left(k\right)}=\boldsymbol{\Pi}^{\left(k\right)}-\psi_{\mathbf{G}}^{-\left(k\right)}\boldsymbol{\Sigma}^{-\left(k\right)}\boldsymbol{\Pi}^{\left(k\right)}\bar{\mathbf{Y}}_{-1}\bar{\mathbf{M}}\bar{\mathbf{Y}}_{-1}^{T}-\xi\psi_{\mathbf{G}}^{-\left(k\right)}\boldsymbol{\Pi}^{\left(k\right)}\mathrm{diag}\left(\mathbf{q}^{\left(k\right)}\right)+\psi_{\mathbf{G}}^{-\left(k\right)}\mathbf{H}^{\left(k\right)}$.

Finally, the majorized subproblem for problem \eqref{eq:Problem Formulation}
is
\begin{equation}
\begin{array}{c}
\underset{\boldsymbol{\Pi}}{\mathsf{minimize}}\:\left\Vert \boldsymbol{\Pi}-\mathbf{P}^{\left(k\right)}\right\Vert _{F}^{2}\;\mathsf{subject\,to}\:\mathrm{rank}\left(\boldsymbol{\Pi}\right)\leq r.\end{array}\label{eq:subproblem}
\end{equation}
This problem has a closed form solution. Let the singular value decomposition
for $\mathbf{P}$ be $\mathbf{P}=\mathbf{U}\mathbf{S}\mathbf{V}^{T}$,
the optimal $\boldsymbol{\Pi}$ is $\boldsymbol{\Pi}^{\star}=\mathbf{U}\mathbf{S}_{r}\mathbf{V}^{T}$,
where $\mathbf{S}_{r}$ is obtained by thresholding the smallest $\left(P-r\right)$
diagonal elements in $\mathbf{S}$ to be zeros. Accordingly, parameters
$\boldsymbol{\alpha}$ and $\boldsymbol{\beta}$ can be factorized
by $\boldsymbol{\Pi}^{\star}=\boldsymbol{\alpha}^{\star}\boldsymbol{\beta}^{\star T}$.

\subsection{The MM-RSVECM Algorithm}

Based on the MM method, to solve the original problem \eqref{eq:Problem Formulation},
we just need to iteratively solve a low-rank approximation problem
\eqref{eq:subproblem} with a closed form solution at each iteration.
The overall algorithm is summarized in Algorithm \ref{alg:MM-RSVECM}.\vspace{-8bp}

\begin{algorithm}[tbh]
\textbf{Input:} $\left\{ \mathbf{y}_{i}\right\} _{i=1}^{N}$ and needed
pre-sampled values.

\textbf{Initialization:} $\boldsymbol{\Pi}^{\left(0\right)}\left(\boldsymbol{\alpha}^{\left(0\right)},\boldsymbol{\beta}^{\left(0\right)}\right)$,
$\boldsymbol{\Gamma}^{\left(0\right)}$, $\boldsymbol{\Sigma}^{\left(0\right)}$
and $k=1$. 

\textbf{Repeat}
\begin{enumerate}
\item Compute $\mathbf{w}^{\left(k\right)}$, $\mathbf{q}^{\left(k\right)}$,
$\mathbf{G}^{\left(k\right)}$, $\mathbf{H}^{\left(k\right)}$, $\psi_{\mathbf{G}}^{\left(k\right)}$
and $\mathbf{P}^{\left(k\right)}$;
\item Update $\boldsymbol{\Pi}^{\left(k\right)}$ by solving \eqref{eq:subproblem}
and $\boldsymbol{\Gamma}^{\left(k\right)}$, $\boldsymbol{\Sigma}^{\left(k\right)}$;
\item $k=k+1$;
\end{enumerate}
\textbf{Until} $\boldsymbol{\Pi}^{\left(k\right)}$, $\boldsymbol{\Gamma}^{\left(k\right)}$
and $\boldsymbol{\Sigma}^{\left(k\right)}$ satisfy a termination
criterion.

\textbf{Output:} $\hat{\boldsymbol{\Pi}}\left(\hat{\boldsymbol{\alpha}},\hat{\boldsymbol{\beta}}\right)$,
$\hat{\boldsymbol{\Gamma}}$ and $\hat{\boldsymbol{\Sigma}}$.\caption{\label{alg:MM-RSVECM}MM-RSVECM - Robust MLE of Sparse VECM}
\end{algorithm}
\vspace{-15bp}

\section{Numerical Simulations\label{sec:Numerical-Simulations}}

\begin{figure}[H]
\begin{centering}
\includegraphics[scale=0.43]{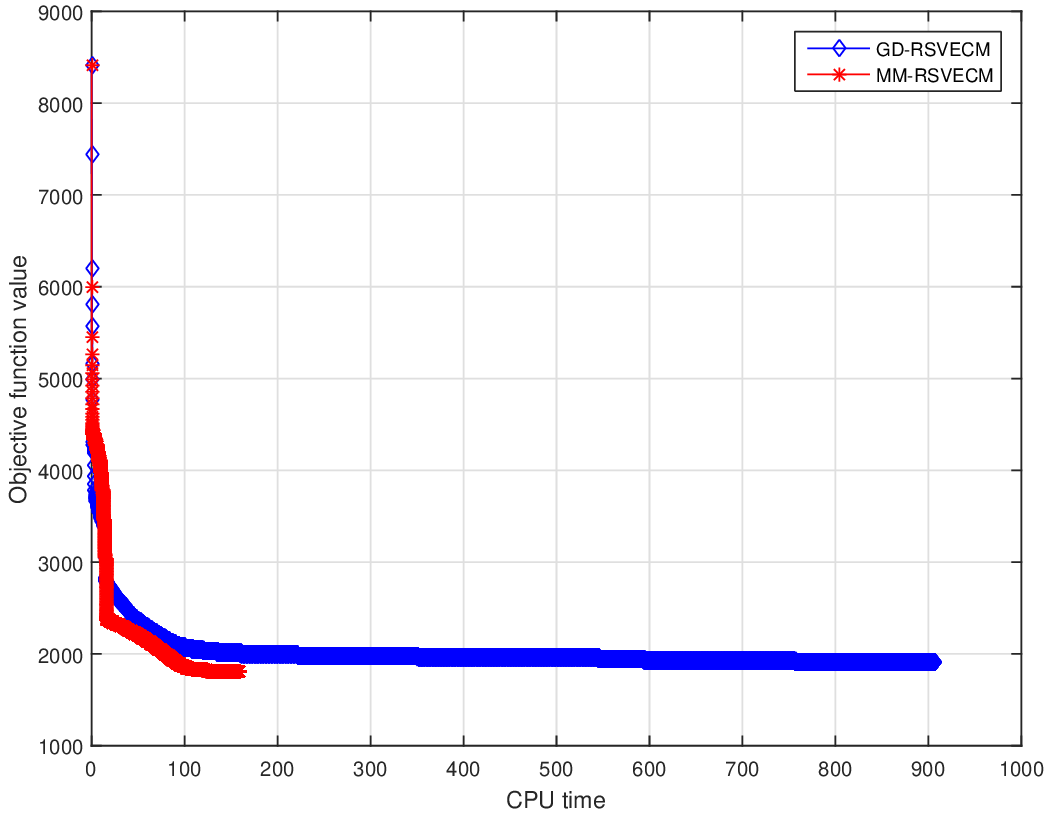}
\par\end{centering}
\centering{}\caption{\label{fig:Convergence-comparison-for}Convergence comparison for
objective function value.}
\end{figure}

Numerical simulations are considered in this section. A VECM $\left(K=5,\:r=3,\:N=1000\right)$
with underlying group sparse structure for $\boldsymbol{\Pi}$ is
specified firstly. Then a time series sample path is generated with
innovations distributed to Student $t$-distribution with degree of
freedom $p=3$. We first compare our algorithm (MM-RSVECM) with the
gradient descent algorithm (GD-RSVECM) for the proposed nonconvex
problem formulation in \eqref{eq:Problem Formulation}. The convergence
result of the objective function value is shown in Fig. \ref{fig:Convergence-comparison-for}.

Based on the MM method, MM-RSVECM obtains a faster convergence than
GD-RSVECM. This may be because the algorithm based on the MM method
better exploits the structure of the original problem. 

\begin{figure}[H]
\begin{centering}
\includegraphics[scale=0.43]{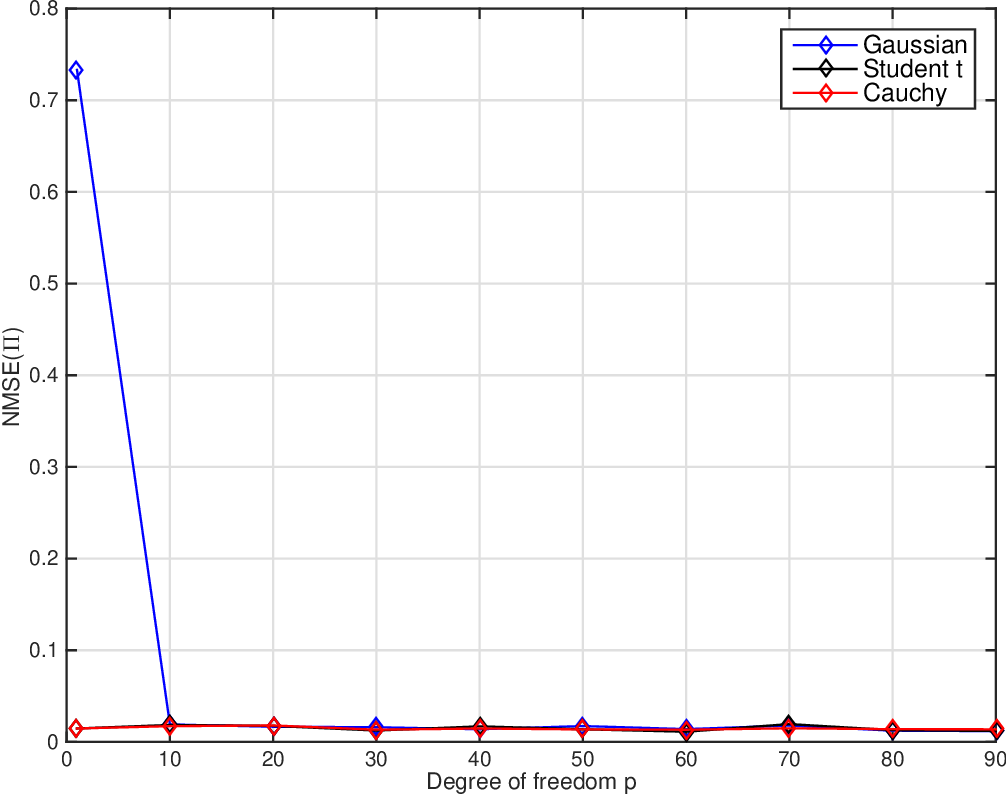}
\par\end{centering}
\caption{\label{fig:NMSE-vs-degree}$\text{NMSE}\left(\boldsymbol{\Pi}\right)$
vs degree of freedom $p$ for $t$-distributions.}
\end{figure}

Then the proposed problem formulation based on Cauchy log-likelihood
loss function is further validated by comparing the parameter estimation
accuracy under student $t$-distributions with different degree of
freedom $p$. The estimation accuracy is measure by the normalized
mean squared error (NMSE):
\[
\text{NMSE}\left(\boldsymbol{\Pi}\right)=\frac{\mathbb{E}\left[\left\Vert \hat{\boldsymbol{\Pi}}-\boldsymbol{\Pi}_{\mathrm{true}}\right\Vert _{F}^{2}\right]}{\left\Vert \boldsymbol{\Pi}_{\mathrm{true}}\right\Vert _{F}^{2}}.
\]
In Fig. \ref{fig:NMSE-vs-degree}, we show the simulation results
for $\text{NMSE}\left(\boldsymbol{\Pi}\right)$ by using three estimation
methods, which are based on Gaussian innovation assumption, true Student
$t$-distribution, and the proposed Cauchy innovation assumption.

From Fig. \ref{fig:NMSE-vs-degree}, we can see the parameter estimated
from Cauchy assumption using the MM-VECM algorithm consistently has
a lower parameter estimation error compared to the estimation results
from Gaussian assumption and even using the true Student $t$-distribution.
Based on this, the proposed problem formulation is validated.

\section{Conclusions\label{sec:Conclusions}}

This paper has considered the robust and sparse VECM estimation problem.
The problem has been formulated by considering a robust Cauchy log-likelihood
loss function and a nonconvex group sparsity regularizer. An efficient
algorithm based on the MM method has been proposed with the efficiency
of the algorithm and the estimation accuracy validated through numerical
simulations.

\pagebreak{}

 \bibliographystyle{IEEEtran}
\bibliography{/Users/ziping/Dropbox/Research/Reference/RefAll}

\end{document}